\documentclass[conference,compsoc]{IEEEtran}
\ifCLASSOPTIONcompsoc
  \usepackage[nocompress]{cite}
\else
  \usepackage{cite}
\fi
\ifCLASSINFOpdf
\else
\fi
\usepackage{tipa}
\usepackage{phonrule}
\usepackage{placeins}
\usepackage{amsmath}
\usepackage{hyperref}           
\usepackage[ruled,vlined]{algorithm2e}
\usepackage{caption} 
\usepackage{graphicx}
\usepackage{algpseudocode}

\AtBeginDocument{%
  \providecommand\BibTeX{{%
    \normalfont B\kern-0.5em{\scshape i\kern-0.25em b}\kern-0.8em\TeX}}}

\begin{document}
%
\title{Markov Chain Monte-Carlo Phylogenetic Inference Construction in Computational Historical Linguistics}

\author{\IEEEauthorblockN{Tianyi Ni}
\IEEEauthorblockA{Department of English\\
Arizona State University\\
Tempe, Arizona 85281\\
Email: tianyin1@asu.edu}
}


%


\maketitle

\begin{abstract}
More and more languages in the world are under study nowadays, as a result, the traditional way of historical linguistics study is facing some challenges. For example, the linguistic comparative research among languages needs manual annotation, which becomes more and more impossible with the increasing amount of language data coming out all around the world. Although it could hardly replace linguists work, the automatic computational methods have been taken into consideration and it can help people reduce their workload. One of the most important work in historical linguistics is word comparison from different languages and find the cognate words for them, which means people try to figure out if the two languages are related to each other or not. In this paper, I am going to use computational method to cluster the languages and use Markov Chain Monte Carlo (MCMC) method to build the language typology relationship tree based on the clusters.
\end{abstract}


%
\IEEEpeerreviewmaketitle

\section{Introduction}
\begin{figure}
    \centering
    \includegraphics[width=0.5\textwidth]{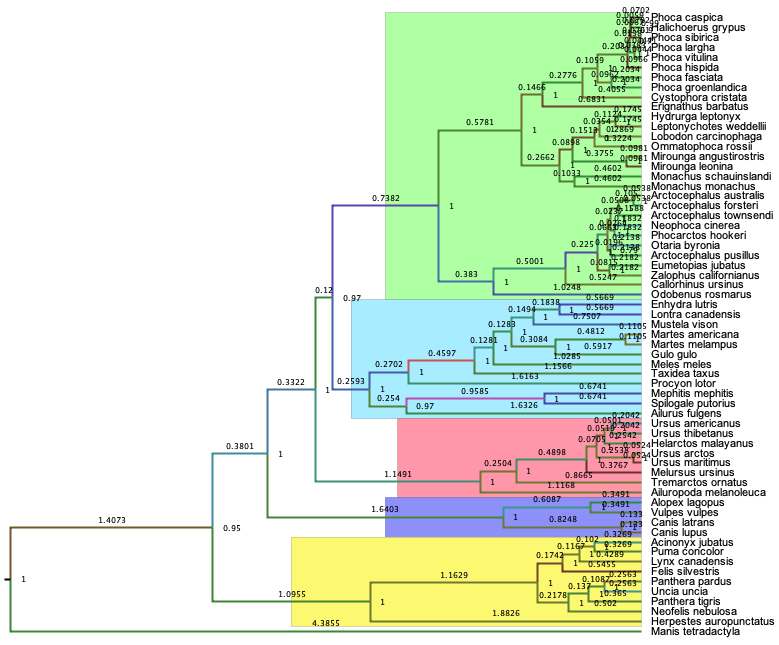}
    \caption{\label{fig:my_label3} The evolution phylogenetic tree of the carnivores. The data in \cite{treegene} are used to generate the tree.}
    
\end{figure}

Bayesian inference of phylogeny has great impact on evolutionary biology. It is believed that all the species are related through a history of a common descent \cite{Huelsenbeck2001}, that is to say, the reason we have various wildlife, including human beings, is because of evolution. We can show the process of evolution and solve the problems, like what is the phylogeny of life, by showing a phylogenetic tree (see Figure \ref{fig:my_label3}). 
As a matter of fact, any research of DNA sequences or protein pattern taken from the species or an individual wildlife can start with a phylogenetic analysis. Language change, which is regarded as just like life evolution, can similarly borrow the phylogenetic analysis to discover the process in which how languages change. It is accepted that all human languages start from the proto-language which is the ancestor of modern languages. For example, Germanic languages (like English, German, Dutch), Romance languages (like Spanish, French and Portuguese), Slavic languages (like Russian, Bosnian and Polish) are from Indo-European languages, which was developed from proto-Indo-European languages. Therefore, we can borrow the computational method from biology and evolution to apply to the language topology.\\
Not only Natural Language Processing (NLP), more and more linguistic branch disciplines begin to have digital datasets and use computational methods to solve some problems. Historical linguistics recently have dramatically increasing digital datasets. The availability of new data and researches of different language from different language families start to challenge the traditional way to carry on the study of historical linguistics. The comparative method has been the core method for linguistic reconstruction for the past 200 years, and is based on manually identifying systematic phonetic correspondences between many words in pair of languages \cite{1}, because different culture has different writing system and different way to spell their words. That is why International Phonetic Alphabet was created to help linguists analyze different languages in parallel. However, there are too few scholars, i.e., historical linguists, to analyze the world's over 7500 type of languages \cite{1,2}, including thousands of languages that have not been studied and are facing extinction. Thereafter, computational methods can help people do researches on unstudied languages faster and give a possible solution. Phylogenetic inference of human languages task is composed with two parts: \textbf{cognate set detection} and \textbf{phylogenetic tree construction}. Cognate set detection automatically assists people put language words with similar or possible evolutionary patterns to one cluster. The phylogenetic tree construction task build trees given the information from the clusters.
In the following, I will divided the whole paper into two main steps: the way I implement cognate detection would be discussed in section 2. After that, I will use the cluster data to carry on the phylogenetic inference program and build the relationship trees, which I will describe the way that how I finished this part in section 3. I will show the results and evaluate them in section 4, and make a conclusion in the last section 5.

\section{Cognate Detection}
A great number of algorithms and mechanisms \footnote{All the algorithms discussed here are applied in LingPy, which is a python3 historical linguistics package. It is used in this project.} to antomatic cognate detection which could be applied in historical linguistics have been used and tested if they are working by many linguists and computer scientists \cite{1,3,4,8,11}. In detail, many of these works are very similar to each other, which consist of two main stages. For the first stage, they first extract the words with same meaning from the wordlists of different languages, either same or different language families, and compare them and use the distance calculation matrix to compute how similar they are. Regarding the second stage, a flat cluster algorithm or a network partitioning algorithm is used to partition all words into cognate sets, and also take the information in the matrix of word pairs as basis \cite{3, 4}. However, the methods in which those researchers use to compare the word pairs are totally different in that people could use different methods to pre-process their language datasets, or even use different algorithms to finish the comparison and clustering task. For example, intuitively, people will start the automated word comparison by computing the distance between the words, such as word embedding in NLP, GloVe \cite{5}, which computes the semantic similarities between the two words. In computational historical linguistics, phonetic segments are used to calculate how close the two words are instead, because the semantics of a single word is not changing as easy as phonetics is. The problem is since the program involves the data pre-processing, then the whole dataset would be traverse twice and the computation would be a big problem when the dataset is about to be over 100 languages. Consequently, people began to think about a faster method to help.
\begin{table}[]
    \centering
    \caption{Some examples of consonants in IPA, also used in the experiments. \label{tab:table1}}
    \begin{tabular}{||c | c||}
    \hline
        Consonant type & Int'l Phoneic Alphabet (IPA)\\
        \hline \hline
        velars & \textbf{k, g, x} \\
        dentals & \textbf{t, d, \textbaro}\\
        liquids & \textbf{r, l, \textinvscr}\\
        nasals & \textbf{n, m, \textltailm, \textscn}\\
        \hline
    \end{tabular}
\end{table}
\subsection{Consonant Class Matching Algorithm (CCM)}
The first linear time method was proposed by \cite{6}, and later modified by \cite{8}. The algorithm compares the word pairs by their \textit{consonant class}. A consonant class is hereby understood as a rough partitioning of speech sounds into groups that are conveniently used by historical linguists when comparing languages \cite{3}. Table \ref{tab:table1} shows some of the international phonetic alphabet (IPA) (for more details about IPA, please refer to \cite{7}). After getting the IPA of the word pairs from the wordlist, the algorithm is to determine if they are cognate by judging their first two consonants class match each other or not. However, since the algorithm only compares the first two consonant classes, the accuracy of this algorithm is relatively low. I have two reasons for this: \textbf{(a)} In linguistics, the number of possible sounds in human languages in the world, excluding artificial languages, amounts to the thousands \cite{4}. It is unrealistic to enroll all the sounds into the system. If we enroll all the sounds in the algorithm to simulate the language change process, we need to get a new matrix in which the probabilities of one sound switching to the other sound will be calculated, which is very time-consuming. \textbf{(b)} comparing the first two consonant classes are not sufficient to determine if the two words in pair are derived from the same cognate word.
\subsection{Edit Distance} The Edit Distance approach is to take the normalized Levenshtein distance \cite{9}, which is a concept in information theory. It aims to measure the difference of two sequences by calculating the minimum number of character or string edits, such as insertion, deletion, which are coincidentally two basic language phonetic change. The distance could be used as a probability to estimate how possible one word changes to the other one.
\subsection{Sound Class Algorithm (SCA)}
This algorithm is for pairwise and multiple alignment analysis \cite{1}. It not only takes an expanded sound class into account, but it considers the prosodic aspect of each word. As a result, it is able to align within morpheme boundaries instead of the sound segments, suppose the morpheme information is the prior knowledge and we already have it.
\subsection{LexStat}
The previous three methods use the same strategy to put the words from different languages into clusters, i.e., UPGMA clustering algorithm, while LexStat uses language-specific scoring schemes which are derived from a Monte-Carlo permutation of the data \cite{4}. The word pairs from different languages are first aligned and scored, and the MC permutation shuffled the word pairs according to their scores. The scores could be calculated by the frequencies of daily use by native speakers. Thus, a distribution of sound-correspondence frequencies is generated. Then, the distribution is used to compare with an attested distribution and then converted into a language-specific scoring scheme for all word pairs.\\
Following the algorithms above, with the consideration of both the advantages and disadvantages of them, in this project, I am going to use a modified method: \textit{sound-class based skip-grams with bipartite networks (BipSkip)}. The whole procedure is quite straightforward and could be divided into three steps. \textbf{First step:} the word pair and their skip-grams are two sets of the bipartite networks. The \textbf{second step} is optional, which is to refine the bipartite network. Before I run the program, I will be asked to input a threshold, which determines if the program should delete the skip-gram nodes linked to fewer word nodes than the threshold itself. According to the experiment, even though I did not input any threshold as one of the parameters, the algorithm could still give the same answer but with more executing time. In the \textbf{last step}, the final generated bipartite graph would be connected to a monopartite graph and partitioned into cognate sets with the help of graph partitioning algorithms. Here I would use Informap algorithm \cite{10}. To make a comparison to this method, I am using CCM and SCA for distance measurement in this experiment, too. UPGMA algorithm would be used accordingly in these two cases. 

\section{Bayesian Phylogenetic Inference}
\begin{figure}
    \centering
    \includegraphics[width=0.35\textwidth]{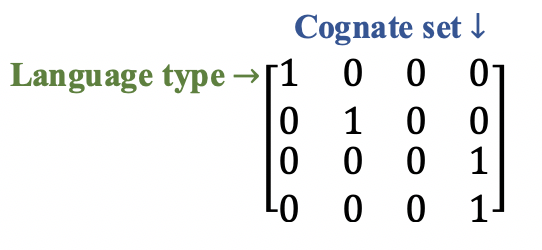}
    \caption{\label{fig:my_label} Data matrix example}
    \vspace{-3mm}
\end{figure}

Methods for Bayesian phylogenetic inference in evolutionary biology and historical linguistics are all based on the following Bayes rule \cite{3, 13}:
\[f(\Lambda |X) = \frac{f(X|\Lambda)f(\Lambda ) }{f(X)}\] or \[Pr(Tree |Data) = \frac{Pr(Data|Tree)\times Pr(Tree)}{Pr(Data)}\]
Where \(f\) means the probability density function, $\Lambda$ consists of the tree topology $\tau$, the branch length vector of the tree $T$ and the substitution model parameter $\theta$; $X$ is a data matrix with dimension $N*K$, within which there are $N$ rows indicating $N$ different words from different kinds of languages and they can be put into $K$ cluster in a language family. 
Figure \ref{fig:my_label} shows an example of the matrix. As we can see, the data matrix is a binary matrix with every element $i_{ij}$. $1$ means language $i$ could be classified into cognate $j$, while $0$ means language $i$ does not belong to cluster $j$. Based on the shape of the data matrix, to get the parameter ($\Lambda = \tau, X, \theta$) for tree generation, we need to sum over the whole matrix and will have to compute all possible topologies of \(\frac{(2N-3)!}{2^{N-2}(N-2)!}\). 

This method is practical only for a small amount of dataset and the calculation is not easy once it is over 5 languages \cite{13}. In addition, based on the Bayesian formula, if we use prior probability $Pr(tree)$ and likelihood $Pr(Data|Tree)$ to produce the posterior probability\footnote{The posterior probability represents how likely the tree is correct. The tree with the largest posterior probability can be treated as the best estimate of the phylogenetic inference tree.}, it seems that the posterior probability is very easy and convenient to formulate based on the Bayesian formula, but to compute this probability and get the greatest estimate of the final tree, the machine has to compute and compare all the possible trees and in each tree, the machine will compute all the combination of branches with different length.\\
Metropolis-Hastings (MH) algorithm \cite{hastings1970} , one of the Markov Chain Monte Carlo techniques, would be a good tool to overcome this computation difficulty by avoiding summing over all of the topologies by evaluating the posterior probability \(f(\Lambda |X)\). The likelihood in this case from one data to the next parameter is calculated by the prunning algorithm. Prunning algorithm, or we can call it K81 model \cite{Felsenstein1981}, is a Markov model of DNA evolution, and this model is a description of the DNA in evolution as a string of four discrete state, i.e., G, C, A, T. Fortunately, language model is very similar to DNA model in that both of them are discrete models , in the language model, we only apply this model in the binary dataset.
\begin{figure}
    \centering
    \includegraphics[width=0.5\textwidth]{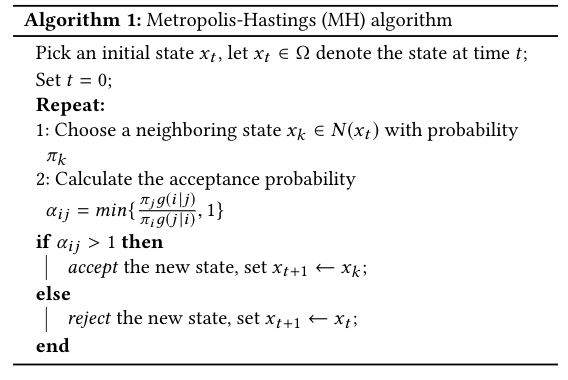}
\end{figure}
\subsection{Markov Chain Monte Carlo (MCMC)}
TBA
\section{Experiments}
\subsection{Settings and Implementation}
It is not easy to figure out which kind of skip-gram and sound-class system would generate the satisfying results, so I design my experiment as follows. \textbf{(1)} During the training process, I would use datasets proposed by \cite{11}, I will give the training dataset in table \ref{tab:my_label6}, and compute skip-gram by length 4. Although there are languages repeated in both training language dataset and testing language dataset, such as Chinese in Sino-Tibet and Austronesian, I manually tick out the repeated data from both dataset and also the concept from both datasets are not overlapping each other.  \textbf{(2)} Regarding sound-class, I would refer to SCA sound class \cite{11} \textbf{(3)} Set the threshold in step 2 as 0.2 (20\%). \textbf{(3)} The evaluation matrix is B-Cubes \cite{12}. The F-score based on the configuration above is, when I use connected components partitioning, \textbf{85.4\%}, and \textbf{85.2\%} when I use Infomap partitioning algorithm.
    \begin{table}[]
    \centering
     \caption{\label{tab:my_label6} Training data from \cite{11}}
    \begin{tabular}{||l c c c||}
    \hline
        \textbf{Dataset} & \textbf{Concepts} & \textbf{Languages} & \textbf{Cognates} \\
        \hline \hline
        Austronesian & 210 & 20 & 2864 \\
        Bai & 110 & 9 & 285 \\
        Chinese & 140 & 15 & 1189 \\
        Japanese & 200 & 10 & 460 \\
        Ob-Ugrian & 110 & 21 & 242 \\
        \hline
    \end{tabular}
\end{table}
\subsection{Evaluation Methods}
TBA
\subsection{Results and Discussion}
\textbf{Cognate Detection} Table \ref{tab:my_label2} shows the statistics in test data developed by \cite{1}. The result of the BipSkip approach for cognate detection is shown in the table \ref{tab:my_label3} as well as the result of using SCA and CCM. As shown in the tables, we can see that the BipSkip approach is not the quickest method, although it is more user-friendly. CCM is surprsingly fastest method with slightly higher precision than BipSkip approach, especially in Austronesian and Sino-Tibetan. Indo-European languages stays almost the same, my guess is because the languages in Indo-European language family are more phonetically similar to each other than other languages in their corresponding language families, as a result of which the three methods would perform almost the same. SCA algorithm is not recommended here in that it costs the longest time and the largest spece since I need to prepare the expanded sound class and also some morphological features.
\begin{table}[h]
    \centering
     \caption{\label{tab:my_label2} Test data from \cite{1}}
    \begin{tabular}{||l c c c||}
    \hline
        \textbf{Dataset} & \textbf{Concepts} & \textbf{Languages} & \textbf{Cognates} \\
        \hline \hline
        Austro-Asiatic & 200 & 58 & 1872 \\
        Austronesian & 210 & 45 & 3804 \\
        Indo-European & 208 & 42 & 2157 \\
        Pama-Nyungan & 183 & 67 & 6634 \\
        Sino-Tibetan & 110 & 64 & 1402 \\
        \hline
    \end{tabular}
\end{table}

\begin{table}[h]
    \centering
    \caption{\label{tab:my_label3} Three methods to extract the cognate on test data.}
    \begin{tabular}{||l l l l||}
    \multicolumn{4}{c}{\textbf{Infomap analysis on test data}}\\
     \hline
        \textbf{Dataset} & \textbf{Precision} & \textbf{Recall} & \textbf{F-score} \\
        \hline \hline
        Austro-Asiatic & 0.69 & 0.78& 0.7323 \\
        Austronesian & 0.71 & 0.74 & 0.7233 \\
        Indo-European & 0.81 & 0.72 & 0.7616 \\
        Pama-Nyungan & 0.53 & 0.69 & 0.6321 \\
        Sino-Tibetan & 0.64 & 0.62 & 0.6033 \\
        \hline
        \textsc{Total} & 0.688 & 0.732 & 0.7066\\
        \hline
        \multicolumn{4}{l}{Running Time: 18.45s} \\
         \multicolumn{4}{l}{} \\
        \multicolumn{4}{c}{\textbf{SCA analysis on test data}}\\
        \hline
        \textbf{Dataset} & \textbf{Precision} & \textbf{Recall} & \textbf{F-score} \\
        \hline \hline
        Austro-Asiatic & 0.72 & 0.80& 0.7601 \\
        Austronesian & 0.82 & 0.74 & 0.7748 \\
        Indo-European & 0.89 & 0.74 & 0.8063 \\
        Pama-Nyungan & 0.59 & 0.85 & 0.6930 \\
        Sino-Tibetan & 0.73 & 0.46 & 0.5614 \\
        \hline
        \textsc{Total} & 0.75 & 0.7180 & 0.7191\\
        \hline
        \multicolumn{4}{l}{Running Time: 240.050s} \\
        \multicolumn{4}{l}{} \\
        \multicolumn{4}{c}{\textbf{CCM analysis on test data}}\\
        \hline
        \textbf{Dataset} & \textbf{Precision} & \textbf{Recall} & \textbf{F-score} \\
        \hline \hline
        Austro-Asiatic & 0.79 & 0.64& 0.7070 \\
        Austronesian & 0.88 & 0.58 & 0.6963 \\
        Indo-European & 0.89 & 0.64 & 0.7484 \\
        Pama-Nyungan & 0.64 & 0.82 & 0.7194 \\
        Sino-Tibetan & 0.78 & 0.35 & 0.4831 \\
        \hline
        \textsc{Total} & 0.7960 & 0.6060 & 0.6708\\
        \hline
        \multicolumn{4}{l}{Running Time: 3.247s}
    \end{tabular}
        \vspace{-3mm}
\end{table}
\textbf{Phylogenetic Inference} 
The result of phylogenetic inference in modified MH algorithm is shown in table \ref{tab:my_label5}. I designed a branch of experiments, changing the settings and get some good results. I set the initial temperature as $T_{0} = \{10, 20, 30, 40, 50, 60, 70, 80, 90, 100\}$. During the project, I will take down the number of iteration, the time it takes to finish running the program. Table \ref{tab:my_label4} is the ground truth results, testing on golden standard cognates from Glottolog. It is hard to determine which one is outperforming the other two. Overall, Indo-European costs the shortest time and fewer iterations, since in the three methods this language family always has the highest the accuracy. In addition, the cognate sets from the automatic approaches is easier to build phylogenetic trees than the manually annotated standard cognate sets, from the result, the automatic methods obviously shorten the time of automatic phylogenetic inference.
\begin{table}[h]
    \centering
        \caption{\label{tab:my_label4} showing the results for golden standard cognates from \cite{3}}
    \begin{tabular}{|| l c c c c ||}
    \hline
        Family & $T_{0}$ & GQD & \#interation & Time \\
        \hline \hline
        Austro-Asiatic & 90 & 0.058 & 26900 & 476.113 \\
        Austronesian & 80 & 0.0389 & 21280 & 123.167 \\
        Indo-European & 10 & 0.0135 & 2260 & 16.713 \\
        Pama-Nyungan & 10 & 0.061 & 22600 & 605.319 \\
        Sino-Tibetan & 50 & 0.0475 & 25700 & 206.952 \\
        \hline
    \end{tabular}
\end{table}
\begin{table}[]
    \centering
    \caption{\label{tab:my_label5} Three methods to extract the cognate on test data.}
    \begin{tabular}{||l c c c c||}
    \multicolumn{4}{c}{\textbf{Test on Infomap cognate}}\\
     \hline
        Family & $T_{0}$ & GQD & \#interation & Time \\
        \hline \hline
        Austro-Asiatic & 80 & 0.0245& 21248 & 310.403\\
        Austronesian & 10 & 0.0875 & 9040 & 82.443\\
        Indo-European & 100 & 0.0690 &2710 & 28.691\\
        Pama-Nyungan & 70 & 0.0159 & 21120 & 662.447\\
        Sino-Tibetan & 80 & 0.3268 & 10640 & 129.903\\
        \hline
         \multicolumn{4}{l}{} \\
        \multicolumn{4}{c}{\textbf{Test on SCA cognate}}\\
        \hline
        Family & $T_{0}$ & GQD & \#interation & Time \\
        \hline \hline
        Austro-Asiatic & 60 & 0.0568& 26523 & 340.605\\
        Austronesian & 10 & 0.0356 & 8560 & 74.230\\
        Indo-European & 10 & 0.469 &3205 & 35.691\\
        Pama-Nyungan & 80 & 0.0658 & 20136 & 302.657\\
        Sino-Tibetan & 50 & 0.2033 & 20365 & 263.219\\
        \hline
        \multicolumn{4}{l}{} \\
        \multicolumn{4}{c}{\textbf{Test on CCM cognate}}\\
        \hline
        Family & $T_{0}$ & GQD & \#interation & Time \\
        \hline \hline
        Austro-Asiatic & 100 & 0.056& 23580 & 310.403\\
        Austronesian & 80 & 0.0785 & 23650 & 309.653\\
        Indo-European & 10 & 0.203 &2510 & 26.723\\
        Pama-Nyungan & 70 & 0.0023 & 25640 & 369.002\\
        Sino-Tibetan & 50 & 0.0569 & 10520 & 185.367\\
        \hline
    \end{tabular}
\end{table}\\
    \vspace{-3mm}
\section{Conclusion}
Obviously, the result is not surprisingly good. We can observe from the data, the accuracy for each language, some of them are only slightly over 50\%. Among the five language families in the testing data, Indo-european has more accuracy than the other four language families, due to the similar phonetic features. Also, some places where native people used to live and was conquered by immigrates, for example languages in the islands in south pacific, Hawaii, Indonesia, etc, their accuracy is obviously high and easy to cluster by cognate and find their relationship. Some native languages in Australia, Pama-Nyungan language family whose languages are mainly on Autralian continent is surprisingly lower than any other southern pacific islands languages.\\
From this exmperiment, we can obviously use this method to help historcial linguists to make an analysis of language development and change, but the result is not very accurate basically. How do we solve the problem? The current datasets only includes the main class of phonetic alphabet, I think it is necessary to enroll some language phonetic change background knowledge to let the machine recognize the probability of change from phoneme $A$ to $B$, such as \textbf{G}reat \textbf{V}owel \textbf{S}hift, etc.
\FloatBarrier


\ifCLASSOPTIONcompsoc



%

\onecolumn

\section*{Appendix}
In this appendix, I am going to show the part of resulting tree generated from Indo-european language testing dataset. The whole tree structure is very big which involves 34 trees totally. I am only showning the first four trees. The number on the edge is the probability that they are related. The labels at the end of the tree are language type: such as \textbf{poli} represents polish, \textbf{norw} represents Norwegian; \textbf{lati} represents Latin. The number behind the language type is the index of the word in the wordlist.
\FloatBarrier
\begin{figure}
    \centering
    \includegraphics[width=0.7\textwidth]{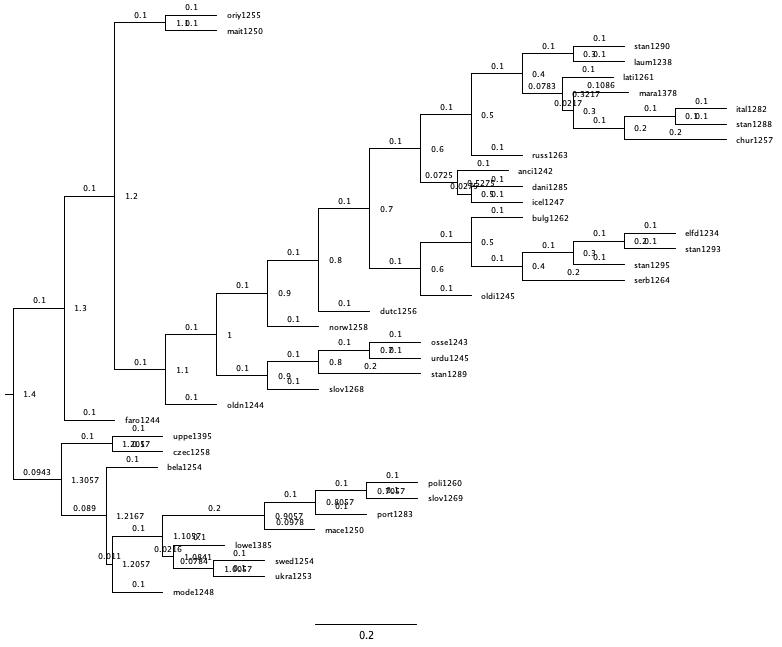}
    
\end{figure}
\begin{figure}
    \centering
    \includegraphics[width=0.7\textwidth]{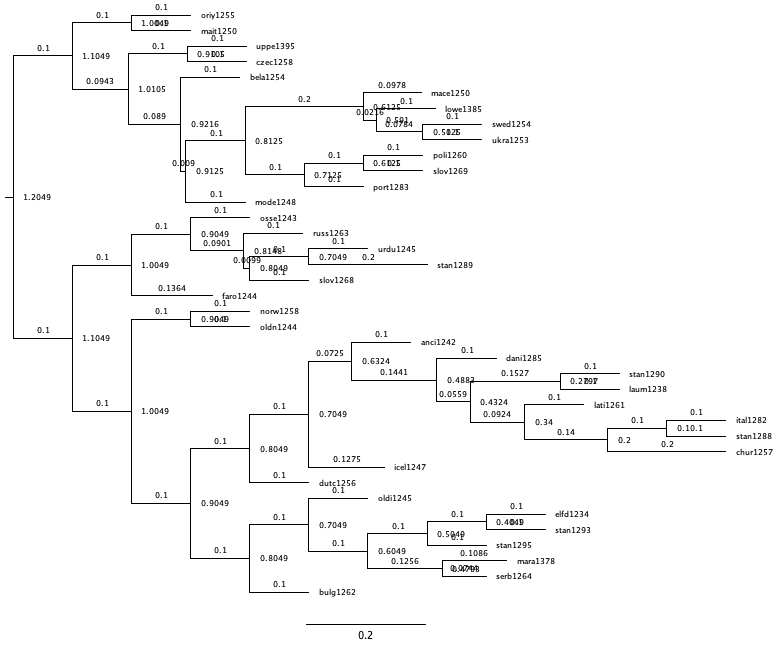}
\end{figure}

\begin{figure}
    \centering
    \includegraphics[width=0.7\textwidth]{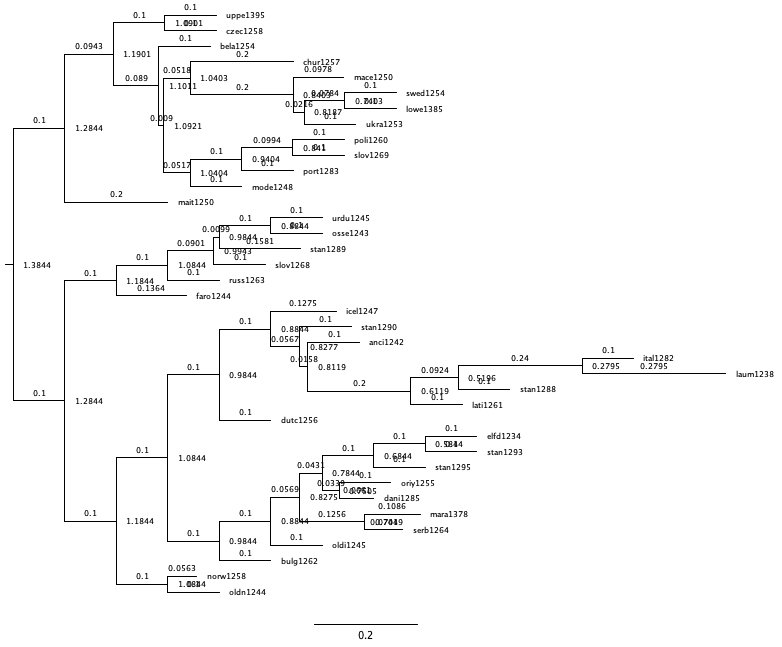}
\end{figure}
\begin{figure}
    \centering
    \includegraphics[width=0.7\textwidth]{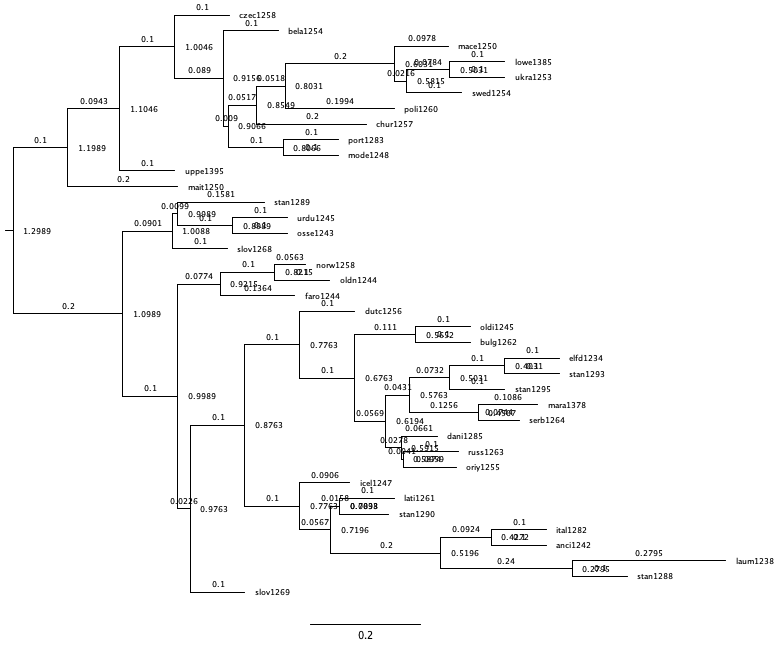}
\end{figure}
\end{document}